%% file: cdc2023.tex
\documentclass[letterpaper, 10 pt, conference]{ieeeconf}

\IEEEoverridecommandlockouts             

\input{settings/packages}

\input{settings/commands}

\usepackage{hologo}
\usepackage{listings}
\begin{document}
\bstctlcite{IEEEexample:BSTcontrol}

\twocolumn[
\begin{@twocolumnfalse}
\Huge {IEEE copyright notice} \\ \\
\large {\copyright\ 2024 IEEE. Personal use of this material is permitted. Permission from IEEE must be obtained for all other uses, in any current or future media, including reprinting/republishing this material for advertising or promotional purposes, creating new collective works, for resale or redistribution to servers or lists, or reuse of any copyrighted component of this work in other works.} \\ \\

{\Large Published in \emph{2023 62nd IEEE IEEE Conference on Decision and Control (CDC)}, Singapore, Singapore, December 13 - 15, 2023.} \\ \\

Cite as:

\vspace{0.1cm}
\noindent\fbox{%
	\parbox{\textwidth}{%
		L.~Ullrich, A.~V{\"o}lz, and K.~Graichen, `Robust Meta-Learning of Vehicle Yaw Rate Dynamics \\ via Conditional Neural Processes,''
		in \emph{2023 62nd IEEE IEEE Conference on Decision and Control (CDC)}, Singapore, Singapore, 2023, pp. 322--327, doi: 10.1109/CDC49753.2023.10384159.
	}%
}
\vspace{2cm}
	
\end{@twocolumnfalse}
]

\noindent\begin{minipage}{\textwidth}
	
\hologo{BibTeX}:
\footnotesize
\begin{lstlisting}[frame=single]
@inproceedings{ullrich2023robust,
	author={Ullrich, Lars and V{\"o}lz, Andreas and Graichen, Knut},
	booktitle={2023 62nd IEEE Conference on Decision and Control (CDC)},
	title={Robust Meta-Learning of Vehicle Yaw Rate Dynamics via Conditional Neural Processes},
	address={Singapore, Singapore},
	year={2023},
	pages={322--327},
	doi={10.1109/CDC49753.2023.10384159},
	publisher={IEEE}
}
\end{lstlisting}
\end{minipage}
\setcounter{page}{0}

\maketitle
\thispagestyle{plain}
\pagestyle{plain}

\input{src/00_Abstract}
\input{src/01_Introduction}
\input{src/02_PhysicalModels}
\input{src/03_ConditionalNeuralProcesses}
\input{src/04_Evaluation}
\input{src/05_Conclusion}

\bibliographystyle{IEEEtran}
\bibliography{literature}

\end{document}

%% file: settings/packages.tex
\usepackage[T1]{fontenc} 
\usepackage[utf8]{inputenc} 
\usepackage[english]{babel} 

\usepackage{amsmath} 
\usepackage{amsfonts}
\usepackage{amssymb}  

\usepackage{booktabs} 

\usepackage{dsfont}
\usepackage{graphicx}
\usepackage{caption}
\usepackage{subcaption}

\usepackage[dvipsnames]{xcolor}
\usepackage{tikz}
\usepackage{pgfplots}
\pgfplotsset{compat=newest}
\usepgfplotslibrary{fillbetween}

\usepackage{epsfig} 
\usepackage{times} 
\usepackage{siunitx}
\usepackage{calc}
\usepackage[hyphens]{url}
\usepackage{svg}
\renewcommand{\baselinestretch}{0.976}

\usepackage{svg}

\usepackage{scalerel}
\usepackage{tikz}
\usetikzlibrary{svg.path}

\definecolor{orcidlogocol}{HTML}{A6CE39}
\tikzset{
	orcidlogo/.pic={
		\fill[orcidlogocol] svg{M256,128c0,70.7-57.3,128-128,128C57.3,256,0,198.7,0,128C0,57.3,57.3,0,128,0C198.7,0,256,57.3,256,128z};
		\fill[white] svg{M86.3,186.2H70.9V79.1h15.4v48.4V186.2z}
		svg{M108.9,79.1h41.6c39.6,0,57,28.3,57,53.6c0,27.5-21.5,53.6-56.8,53.6h-41.8V79.1z M124.3,172.4h24.5c34.9,0,42.9-26.5,42.9-39.7c0-21.5-13.7-39.7-43.7-39.7h-23.7V172.4z}
		svg{M88.7,56.8c0,5.5-4.5,10.1-10.1,10.1c-5.6,0-10.1-4.6-10.1-10.1c0-5.6,4.5-10.1,10.1-10.1C84.2,46.7,88.7,51.3,88.7,56.8z};
	}
}

\newcommand\orcidicon[1]{\href{https://orcid.org/#1}{\mbox{\scalerel*{
				\begin{tikzpicture}[yscale=-1,transform shape]
					\pic{orcidlogo};
				\end{tikzpicture}
			}{|}}}}

\usepackage{hyperref} 
\hypersetup{
	colorlinks=true,
	breaklinks=true,
	linkcolor=blue,
	filecolor=magenta,      
	hypertexnames=true,
	urlcolor=cyan,
	pdfpagemode=FullScreen,
	citecolor=green
}
\usepackage{cleveref}

%% file: settings/commands.tex
\title{\LARGE \textbf{Robust Meta-Learning of Vehicle Yaw Rate Dynamics \\ via Conditional Neural Processes}$^{*}$

\author{Lars Ullrich\textsuperscript{\orcidicon{0009-0001-8166-3118}},~\IEEEmembership{Graduate Student Member,~IEEE}, Andreas Völz\textsuperscript{\orcidicon{0000-0002-8040-6400}}, and Knut Graichen\textsuperscript{\orcidicon{0000-0003-2865-8093}},~\IEEEmembership{Senior Member,~IEEE}
	\thanks{*This research is accomplished within the project ”AUTOtechagil” (FKZ 01IS22088Y). We acknowledge the financial support for the project by the Federal Ministry of Education and Research of Germany (BMBF).}
	\thanks{The authors are with the Chair of Automatic Control, Friedrich-Alexander-Universität Erlangen-Nürnberg (FAU), Germany {\tt\footnotesize \{lars.ullrich, andreas.voelz, knut.graichen\}@fau.de}}%
}}

\definecolor{cblue}{RGB}{0,112,192}
\definecolor{cred}{RGB}{216,80,24}
\definecolor{cgreen}{RGB}{120,184,40}
\definecolor{cyellow}{RGB}{216,170,32}
\definecolor{cpurple}{RGB}{128,48,144}
\definecolor{clightblue}{RGB}{80,192,240}

\makeatletter
\def\bstctlcite{\@ifnextchar[{\@bstctlcite}{\@bstctlcite[@auxout]}}
\def\@bstctlcite[#1]#2{\@bsphack
	\@for\@citeb:=#2\do{%
		\edef\@citeb{\expandafter\@firstofone\@citeb}%
		\if@filesw\immediate\write\csname #1\endcsname{\string\citation{\@citeb}}\fi}%
	\@esphack}
\makeatother

%% file: src/00_Abstract.tex
\begin{abstract}
	Trajectory planners of autonomous vehicles usually rely on physical models to predict the vehicle behavior. However, despite their suitability, physical models have some shortcomings. On the one hand, simple models suffer from larger model errors and more restrictive assumptions. On the other hand, complex models are computationally more demanding and depend on environmental and operational parameters. In each case, the drawbacks can be associated to a certain degree to the physical modeling of the yaw rate dynamics. Therefore, this paper investigates the yaw rate prediction based on conditional neural processes (CNP), a data-driven meta-learning approach, to simultaneously achieve low errors, adequate complexity and robustness to varying parameters. Thus, physical models can be enhanced in a targeted manner to provide accurate and computationally efficient predictions to enable safe planning in autonomous vehicles. High fidelity simulations for a variety of driving scenarios and different types of cars show that CNP makes it possible to employ and transfer knowledge about the yaw rate based on current driving dynamics in a human-like manner, yielding robustness against changing environmental and operational conditions.
\end{abstract}

%% file: src/01_Introduction.tex
\section{INTRODUCTION}
Vehicle models are key components for trajectory planning in highly automated \cite{milanes2013cooperative}, \cite{funke2016collision}, \cite{wen2018cooperative}, \cite{liu2018dynamic} or even autonomous driving \cite{paden2016survey}, \cite{hult2018design}, \cite{betz2019software} that allow to predict the vehicle behavior as a function of the model inputs. Thus, physically feasible trajectories can be generated and evaluated \cite{pek2018computationally}. However, the complexity of the model has a decisive impact. Simple models benefit from lower computational costs but have more restrictive assumptions and a higher model error \cite{kang2014comparative}, \cite{kong2015kinematic}.  For instance, the most simple point-mass model \cite{tomas2013vehicular}, \cite{godbole1997design} abstracts the vehicle to a large extent and therefore neglects the non-holonomic behaviour. Accordingly, already parking maneuver planning requires a more complex model such as the kinematic single-track model \cite{paden2016survey}. For evasive maneuver planning, however, a kinematic model is not sufficient and dynamic single-track models are typically used for this purpose \cite{shiller1991dynamic}, \cite{hwan2011anytime}. When it comes to highly dynamic maneuvers, single-track drift models or even multi-body models are required \cite{chebly2017coupled}, \cite{viana2019cooperative}. 

More sophisticated models reveal a higher level of detail in the tire model and the lateral dynamics. The less restrictive assumptions and reduced systematic model error implies not only a higher complexity and computational effort, but also increases the number of environmental and operating dependent parameters such as friction or vehicle mass \cite{subosits2021impacts}. Since such parameters have a significant impact on driving dynamics, variations of these parameters cannot be neglected in the trajectory planning of highly autonomous vehicles.

One way to improve the robustness to changing conditions is the use of sensor systems \cite{du2019rapid}, \cite{matsuzaki2008rubber}, \cite{lee2017intelligent}, which leads to higher costs and more potential failure sources. Another way might be software-based virtual sensor systems that estimate these parameters from already existing sensor systems \cite{gustafsson2001virtual}, \cite{nakatsuji2007online}. This methodology offers the possibility to create a vehicle model with fewer assumptions and a larger operational domain of validity, which is of particular importance for highly automated driving. 

Numerical investigations of a kinematic single-track model with ground truth yaw angle show that the model is robust to varying parameters and the error is very small even in dynamic situations \cite{zhu2013real}, \cite{lubiniecki2023adaptive}, \cite{kontos2023prediction}. Thus, if the yaw angle could be determined more accurately and robustly without extended modeling of tires and lateral dynamics, prediction speed could be improved. Since the yaw rate is orientation invariant in contrast to the yaw angle, but leads to the angle by integration, the focus is on the modeling of the yaw rate.

An alternative to detailed physical description of the yaw rate is data-driven modeling \cite{jin2019advanced}, \cite{kontos2023prediction}. Likewise, there is the possibility to describe dynamic systems by means of continuous neural ODEs  \cite{rahman2022neural} or to supplement them by Gaussian processes \cite{hewing2019cautious}. Ultimately, the majority of approaches result in one final model. However, vehicle dynamics are subject to different effects depending on the use, thus several specific models and a proper approach for model switching or blending are required for widely accurate predictions \cite{matute2020lateral}. In this area, meta learning could provide a remedy. 

The key principle of meta-learning is learning-to-learn \cite{hospedales2021meta}. This is achieved, for example, by a dual exploitation of the supervised learning methodology. On the meta level, the mapping from a data set to a dedicated predictor is realized. In this paper, the baseline meta-learning approach of conditional neural processes (CNP) \cite{garnelo2018conditional} is used to predict the yaw rate. A detailed review of the work that has been done in the area of neural processes in general is provided by \cite{jha2022neural}. To the best of the authors knowledge, this is the first time that yaw rate predictions are realized a) without detailed physical modeling of tires or lateral dynamics, b) without additional sensors, c) without limiting assumptions on dynamic, environmental, and operational domains, d) thus providing robustness to changing operational and environmental conditions, e) while providing fast predictions through meta-learning. 

The paper is structured as follows: physical models of different complexity are introduced in Section \ref{phys_chap}. The methodology of CNP and the application to yaw rate prediction are described in Section \ref{cnp_chap}. Finally, CNP are evaluated in different scenarios and the results are compared with the various physical models in Section \ref{res_chap}.

%% file: src/02_PhysicalModels.tex
\section{PHYSICAL MODELS}\label{phys_chap} 

Physical models represent the state of the art in modeling yaw rate dynamics and form the basis to compare against CNP. The equations describing the yaw rate of the kinematic single-track model (KST), the dynamic single-track model (DST), and the single-track drift model (STD) are outlined below and represent a concise summary of the benchmark CommonRoad \cite{althoff2017commonroad}.

\subsection{Kinematic Single-Track Model (KST)}
In the case of the simplest physical baseline model, the yaw angle $\Psi$ is modeled directly \cite{althoff2017commonroad}. Therefore, the yaw rate dynamics
\begin{equation}\label{eq:Psi_KST}
	\ddot{\Psi}_{\mathrm{KST}} = \frac{\mathrm{d}}{\mathrm{d}t}\Bigl(\dot{\Psi}\Bigr) = \frac{\mathrm{d}}{\mathrm{d}t} \Bigl(\frac{v}{l_{\mathrm{wb}}} \tan(\delta)\Bigr),
\end{equation}
is determined by the derivative of the yaw angle dynamics $\dot{\Psi}$, where $v$ denotes the velocity, $\delta$ the steering angle, and $l_{\mathrm{wb}}$ the wheelbase.

\subsection{Dynamic Single-Track Model (DST)}

In contrast, the yaw rate dynamics $\ddot{\Psi}_{\mathrm{DST}}$ of the dynamic single-track model \cite{althoff2017commonroad} is given by
\begin{equation}\label{eq:Psi_ST}
	\ddot{\Psi}_{\mathrm{DST}} = \frac{1}{I_{z}} \Bigl(l_{f}C_{f}\delta+(l_{r}C_{r}-l_{f}C_{f})\beta-(l_{f}^{2}C_{f}+l_{r}^{2}C_{r})\frac{\dot{\Psi}}{v}\Bigr)
\end{equation}
with the moment of inertia $I_{z}$ of the vehicle about the $z$-axis, the slip angle  $\beta$ at the center of gravity and the distances $l_{f}$, $l_{r}$ from the center of gravity to the front and rear axle, respectively. The cornering stiffness $C_{f}$, $C_{r}$ for front and rear
\begin{equation}\label{eq:Ci}
	C_{i} = \mu C_{S,i}F_{z,i}, \quad i \in \{f,r\}
\end{equation}
depends on the friction coefficient $\mu$, specific cornering stiffness coefficients $C_{S,f}, C_{S,r}$, and the vertical forces
\begin{equation}\label{eq:Fz}
	F_{z,f} = m\frac{gl_{r}-a_{\mathrm{long}}h_{\mathrm{cg}}}{l_{r}+l_{f}}, \quad F_{z,r} = m\frac{gl_{f}+a_{\mathrm{long}}h_{\mathrm{cg}}}{l_{r}+l_{f}}
\end{equation}
that take into account the load transfer caused by the longitudinal acceleration  $a_{\mathrm{long}}$ as a function of the mass $m$ and the height of the center of gravity $h_{\mathrm{cg}}$. 

\subsection{Single-Track Drift Model (STD)}

The single-track drift model \cite{althoff2017commonroad} extends the consideration of lateral dynamics and the complexity of the tire model. The yaw rate dynamics $\ddot{\Psi}_{\mathrm{STD}}$ of the single-track drift model
\begin{equation}\label{eq:PSI_STD}
	\ddot{\Psi}_{\mathrm{STD}} =\frac{1}{I_{z}} \Bigl(F_{y,f}\cos(\delta)l_{f}-F_{y,r}l_{r}+F_{x,f}\sin(\delta)l_{f}\Bigr)
\end{equation}
is computed by means of the longitudinal tire forces $F_{y,f}, F_{y,r}$ for front and rear as well as the lateral tire force $F_{x,f}$ for front. The longitudinal and lateral tire forces are calculated via the Pacejka magic tire formula for combined slip \cite{pacejka2002tire}. This formula, which is omitted due to space constraints, takes into account the vertical tire forces $F_{z,f}, F_{z,r}$ like in (\ref{eq:Fz}) as well as the lateral tire slip $\alpha_{f}, \alpha_{r}$ for front and rear
\begin{align}\label{eq:alpa}
	\begin{split}
		\alpha_{f} &= \arctan \Bigl( \frac{v\sin(\beta)+\dot{\Psi}l_{f}}{v\cos(\beta)}\Bigr)-\delta\\
		\alpha_{r} &= \arctan \Bigl( \frac{v\sin(\beta)-\dot{\Psi}l_{r}}{v\cos(\beta)}\Bigr),
	\end{split}
\end{align}
and the longitudinal tire slip $s_{f}, s_{r}$ for front and rear
\begin{equation}\label{eq:s}
	s_{f} = 1- \frac{R_{\omega}\omega_{f}}{u_{\omega,f}}, \quad s_{r} = 1- \frac{R_{\omega}\omega_{r}}{u_{\omega,r}},
\end{equation}
with the effective tire radius $R_{\omega}$ and the front and rear tire velocities $u_{\omega,f}, u_{\omega,r}$. The tire velocities can be computed by
\begin{align}\label{eq:u_omega}
	\begin{split}
		u_{\omega,f} &= v\cos(\beta) \cos(\delta) + (v \sin(\beta)+l_{f} \dot{\Psi}) \sin(\delta)\\
		u_{\omega,r} &= v\cos(\beta).
	\end{split}
\end{align}

Since the dynamic single-track model as well as the single-track drift model become singular at low velocities, a distinction is mandatory. In this paper, the simulations  of the physical models are based on the implementation of CommonRoad \cite{althoff2017commonroad}, where the dynamic single-track model is switched for velocities smaller than \num{0.1}  \si{\metre\per\second}. In comparison, the single-track drift model utilizes a more complex model blending \cite{althoff2017commonroad}. The above equations describe only the pure modeling in the cases without singularity issues.

%% file: src/03_ConditionalNeuralProcesses.tex
\section{CONDITIONAL NEURAL PROCESSES (CNP)}\label{cnp_chap}

While supervised learning operates on single datasets of single tasks and provides a task-specific predictor, meta-learning operates on datasets of multiple related tasks and provides an advanced predictor that shares knowledge across tasks to be able to predict even in unseen tasks. 
Since predictions represent a foundation for decision-making in trajectory planning, the consideration of model uncertainty is of critical interest.
The neural process family \cite{jha2022neural} represents a group of models that belong to the Bayesian meta-learning domain and constitute an adequate solution. In the following, the conditional neural processes \cite{garnelo2018conditional} as basic model of this group are described.

\subsection{CNP Framework}\label{3a}
Classical supervised learning approaches have a dataset $\mathcal{D}=\{(\boldsymbol{x}_{n}, \boldsymbol{y}_{n})\}_{n=1}^{N}$ consisting of inputs $\boldsymbol{x}_{n} \in \mathcal{X} \subseteq \mathbb{R}^{d_{\boldsymbol{X}}}$ and outputs $\boldsymbol{y}_{n} \in \mathcal{Y} \subseteq \mathbb{R}^{d_{\boldsymbol{Y}}}$ and learn the input-output mapping function $f: \mathcal{X} \rightarrow \mathcal{Y}$ such that the final weights result in a minimum error over the entire dataset. Thus, a model is trained which approximates the process best on average. However, in the case that the data generating process itself is subject to variations due to changing conditions, the model would neglect them. To counteract this issue, a meta-learning dataset considers a finite set of input-output mapping functions $f: \mathcal{X} \rightarrow \mathcal{Y}$, sampled from a probability distribution $P$ over functions. Each sampled function $f_{j}$ leads to a task dataset $\mathcal{D}_{j}=(\mathcal{C}_{j}, \mathcal{T}_{j})$ containing a labeled context set $\mathcal{C}_{j}=\{(\boldsymbol{x}_{i}, \boldsymbol{y}_{i})\}_{i=1}^{N}$ and an unlabeled target set $\mathcal{T}_{j}=\{\boldsymbol{x}_{i}\}_{i=N+1}^{N+M}$. Thus, the meta-learning dataset $\mathcal{D_{M}}=\{\mathcal{D}_{j}\}_{j=1}^{K}$ emerges as a set of  $K$ task-specific datasets.

The objective of the meta-learning procedure is to capture the stochastic process that generated a given context set $\mathcal{C}_{j}$ to predict the outputs corresponding to the target set  $\mathcal{T}_{j}$. For this purpose, CNP maps the context set into embeddings
\begin{equation}\label{eq:cnp_emb}
	\boldsymbol{e}_{j,i} = \boldsymbol{\phi_{\theta}}(\boldsymbol{x}_{i}, \boldsymbol{y}_{i}) \quad \forall(\boldsymbol{x}_{i}, \boldsymbol{y}_{i}) \in \mathcal{C}_{j}
\end{equation}
using a learnable encoder neural network $\boldsymbol{\phi_{\theta}}: \mathcal{X} \times \mathcal{Y} \rightarrow \mathbb{R}^{d_{\boldsymbol{e}}}$. Here, the dimension of the embeddings ${d_{\boldsymbol{e}}}$ is a design parameter of fixed size. Furthermore, the task-agnostic embedding set $\mathcal{E}_{j}=\{\boldsymbol{e}_{j,i}\}_{i=1}^{N}$ is converted into a single representation
\begin{equation}\label{eq:cnp_agg_2}
	\boldsymbol{e}_{j} = \boldsymbol{a}(\mathcal{E}_{j}) = \boldsymbol{e}_{j,1} \oplus \boldsymbol{e}_{j,2} \oplus \dots \oplus \boldsymbol{e}_{j,N-1} \oplus \boldsymbol{e}_{j,N}
\end{equation}
employing an aggregation function. For better generalization, CNP require that the aggregation function performs commutative operations $\oplus$ to guarantee a single permutation invariant representation $\boldsymbol{e}_{j}$, according to Deep Sets \cite{zaheer2017deep}. In order to consider context datasets of flexible size, the average is usually taken. The encoded context is decoded together with the target set
\begin{equation}\label{eq:cnp_phi}
	\boldsymbol{d}_{j,i} = \boldsymbol{\rho_{\theta}}(\boldsymbol{x}_{i}, \boldsymbol{e}_{j}) \quad \forall\boldsymbol{x}_{i} \in \mathcal{T}_{j}
\end{equation}
by means of a learnable decoder neural network $\boldsymbol{\rho_{\theta}}: \mathcal{X} \times \mathbb{R}^{d_{\boldsymbol{e}}} \rightarrow \mathbb{R}^{d_{\boldsymbol{d}}}$. In regression tasks, the decoding $\boldsymbol{d}_{j,i}$ is used to parameterize the mean $\mu_{j,i}(\boldsymbol{d}_{j,i})$ and variance $\sigma_{j,i}^{2}(\boldsymbol{d}_{j,i})$ of a Gaussian distribution for each $\boldsymbol{x}_{i}$ in $\mathcal{T}_{j}$. The simplest parameterization is the definition of a two-dimensional decoder output ($d_{\boldsymbol{d}}=2$), which directly assigns a value to $\mu$ and $\sigma^{2}$. In addition, CNP are characterized by the fact that multilayer perceptrons (MLP) are used for the encoder-decoder structure.

More generally and compactly, the CNP models the conditional predictive distribution
\begin{equation}\label{eq:cnp_general}
	p(f(\mathcal{T})| \mathcal{T}, \mathcal{C}) = p(f(\mathcal{T})| \boldsymbol{\rho} (\mathcal{T}, \boldsymbol{E}(\mathcal{C}), \boldsymbol{\theta}) 
\end{equation}
for a given context and target set by means of a decoder $\rho$ and an encoding composition $\boldsymbol{E} = \boldsymbol{a} \circ \boldsymbol{\phi} $,  which is a joint distribution over the random variables $f(\boldsymbol{x}_{i})_{i=1}^{N+M}\}$. For regression tasks, the predictive distribution is modeled as a factorized Gaussian over the target set
\begin{equation}\label{eq:cnp_factorized}
	p(f(\mathcal{T})| \mathcal{T}, \mathcal{C}) = \prod_{i=1}^{N+M}  \mathcal{N}(	\mu_{i}, \sigma_{i}^{2}).
\end{equation}
Under this direct parameterization, mathematical guarantees of stochastic processes are traded off in favor of higher flexibility and scalability for conditional predictions, resulting in a complexity of $\mathcal{O}(N+M)$ for $M$ predictions given $N$ context pairs.

\subsection{CNP based Yaw Rate Prediction}

The framework of CNP offers the possibility to consider information of the current driving dynamics in the form of the context set $\mathcal{C}$ and to provide a distribution of predictions $p(f(\mathcal{T})| \mathcal{T}, \mathcal{C})$ depending on this context. Therefore, the dynamics of the yaw rate is interpreted as a stochastic process and the measured data from various circumstances, e.g. different dynamic ranges or weather conditions, as sampled functions $f \thicksim P$ from the distribution $P$ of this stochastic process.

In this paper CarMaker\footnote{\url{https://ipg-automotive.com/en/products-solutions/software/carmaker/}} (CM) is used to generate the meta-learning dataset. The simulation tool CarMaker is widely used in the automotive industry due to its high-quality multi-body model and tire models \cite{ipg1}, \cite{tong2020overview}. Data are selected in two urban cases, two interurban cases, two longitudinal dynamic cases and fourteen lateral dynamic cases to yield a balanced data set, which is a general requirement for accurate predictions \cite{mirus2020importance}. All twenty cases are simulated in different weather conditions, simulated by different friction values ($\mu_{\mathrm{dry}}=1.0, \mu_{\mathrm{wet}}=0.5, \mu_{\mathrm{icy}}=0.2$). Furthermore, the velocity is varied between \SI{0} and \SI{120}{\kilo\meter\per\hour}. As with real-world data, time series of different lengths occur depending on the test run. The used model builds up on the neural process family implementation\footnote{\url{http://yanndubs.github.io/Neural-Process-Family/}.} and extends the framework to use datasets of different length without any additional data selection procedure.

\begin{figure}[thpb]
	\centering	
	\includegraphics[scale=0.3]{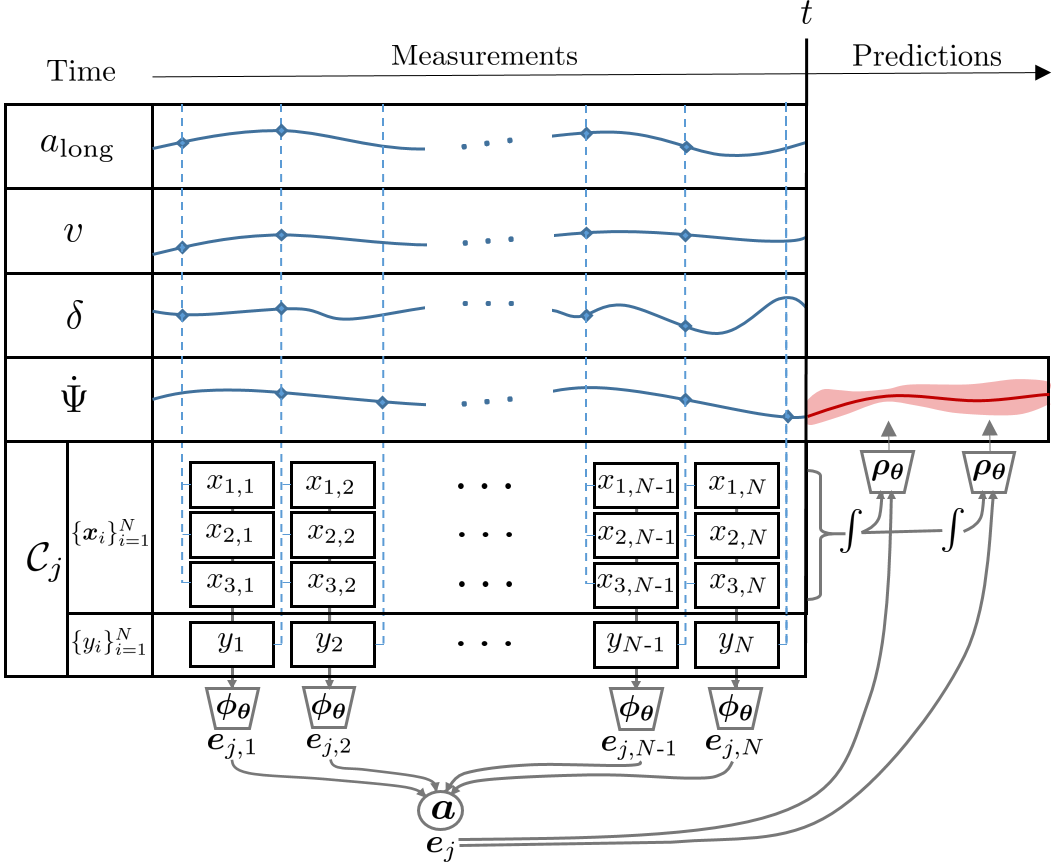}
	\caption{CNP for yaw rate predictions.}
	\label{fig:model}
\end{figure}

An overview of the architecture of the used CNP is listed in the Tab. \ref{table_arc}. There are two encoders listed, which represent the single encoder in Sec. \ref{3a}.  Here, the inputs are encoded first which are further encoded with the targets in the second encoder. As the architecture reveals, an embedding size ${d_{\boldsymbol{e}}}$ of $64$ is used.
\begin{table}[h]
	\caption{CNP Architecture.} 
	\begin{center}
		\begin{tabular}{|c|c|c|c|c|}
			\hline
			General & Network & Hidden & Hidden & Activation\\
			Modules & Architecture & Layers & Size & Function\\
			\hline
			\hline
			Feature Encoder & MLP & 1 & 64 & ReLU\\
			Context Encoder & MLP & 2 & 128 & ReLU \\
			Decoder & MLP & 4 & 64 & ReLU \\
			\hline
		\end{tabular}
		\label{table_arc}
	\end{center}
\end{table}
Inspired by the physical models, different input vectors $\boldsymbol{x}_{i}$ have been investigated for the prediction of the scalar output variable $y_{i}=\dot{\Psi}$. As a result, an input vector consisting of  the steering angle $\delta$, velocity $v$ and the longitudinal acceleration $a_{\mathrm{long}}$ has emerged. The CNP is trained by randomly sampling task data sets from the metadata set, selecting context and target splits, passing the data through the architecture, calculating the loss, and performing a stochastic gradient update until it converges. As a loss function, the negative conditional log likelihood  
\begin{equation}\label{eq:cnp_loss}
	\mathcal{L}(\boldsymbol{\theta}) = - \mathbb{E}_{f \thicksim P} \Bigl[  \mathbb{E}_{S} [ \log p_{\boldsymbol{\theta}}(\{\boldsymbol{y}_{i}\}_{i=1}^{S} | \{\boldsymbol{x}_{i}\}_{i=1}^{S}, \mathcal{C}_{S} )]\Bigr]
\end{equation}
is minimized, where $S$ denotes a subset of the sampled task-set. Since the CNP is trained offline, no gradient updates are necessary during the deployment, which enables fast predictions. The overall deployment workflow of the trained model is outlined in Fig. \ref{fig:model}.

The learning-to-learn approach makes it possible to transfer the knowledge to previously unseen cases and thus offers a computationally efficient alternative to online learning. Due to the strict separation of learning and deployment, no improvement or adaptation can be achieved online, but at the same time catastrophic forgetting \cite{robins1995catastrophic} is counteracted by having a temporally unchangeable system.

%% file: src/04_Evaluation.tex
\section{EVALUATION RESULTS} \label{res_chap}
The methodology is evaluated on the basis of the wide range of scenarios that are also used for the training, but under different environmental and usage conditions. Due to modified conditions, these scenarios represent new samples of the underlying stochastic processes.

In order to investigate the adaptation to changing environmental conditions, variations of the friction coefficient are considered to simulate different road and weather conditions. With regard to changed operating conditions, a changed load (additional four persons, each 80 kg) is selected. For the purposes of generalization, further scenarios as well as the transfer to other vehicles are considered.  The evaluation compares the CNP against the physical models, which are parameterized using the parameters of the CarMaker simulation software, which is used as reference.

\subsection{Friction variation}

\begin{figure}[thpb]
	\input{figures/04_Evaluation_Friction}
	\caption{Evaluation under changed friction coefficients.}
	\label{fig:mue}
\end{figure}
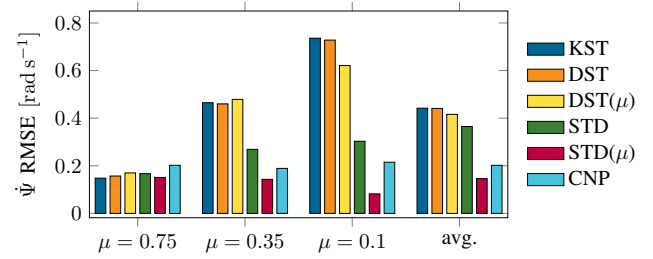

The model is trained using data from 20 scenarios, each at different speeds and for friction coefficients 1.0, 0.5 and 0.2. In this evaluation, the trained model is executed in the same 20 scenarios, each across different velocities for friction coefficients 0.75, 0.35 and 0.1. The context set is based on measurements over a length of \SI{10}{\percent} of each scenario. The target set is determined by Euler integration of the control variables over the length of the scenario. The CNP model is compared against the physical models, which also employ Euler integration for yaw rate prediction. We simulate on the one hand models without online parameter adaption (KST, DST, STD) and models with an ideal online parameter update (DST($\mathrm{\mu}$), STD($\mathrm{\mu}$)). The results in Fig. \ref{fig:mue} show that the physical models perform slightly better than the CNP at a friction coefficient close to dry roads. However, the CNP performs better at lower friction coefficients. The DST model does not benefit from the knowledge of the friction coefficient, but the STD model does. Presumably, this can be attributed to drift behavior of the vehicle in highly dynamic scenarios. The CNP demonstrates consistent performance over the different friction coefficients. On average, only the ideal STD($\mathrm{\mu}$) is superior, which has full knowledge about the current friction coefficient.

\subsection{Mass Variation}
\begin{figure}[thpb]
	\input{figures/04_Evaluation_Mass}
	\caption{Evaluation of prediction under varied mass. Results separated according to different road (friction) conditions.}
	\label{fig:mass}
\end{figure}
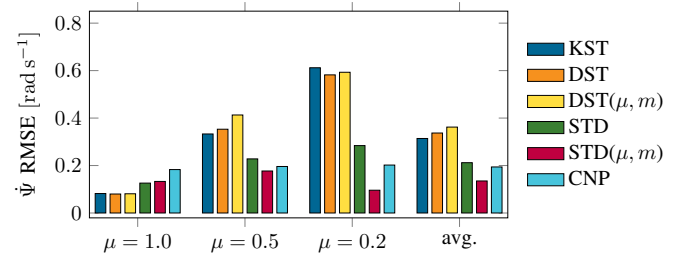

The effect of changed mass is evaluated similarly to the effect of changed friction. The mass was increased for this purpose, the friction remains the same as in the training. The ideal physical models (DST($\mathrm{\mu, m}$), STD($\mathrm{\mu, m}$)) use ground truth mass and friction values in this evaluation. Figure \ref{fig:mass} shows that in the case of dry roads, simple models are slightly better. In the wet and icy road case, more complex models indicate significant advantages. Overall, it can be seen that mass is a less influential factor than friction, but this may be due to uniform load distribution. Again, it becomes apparent that only the ideal STD($\mathrm{\mu,m}$) can benefit from the actual ground truth values. Similar to the first evaluation, the error of the CNP is relatively constant over the variations compared to other models. Again, on average only the STD($\mathrm{\mu,m}$) model performs comparably. 

\subsection{Scenario variation}\label{Scenario}
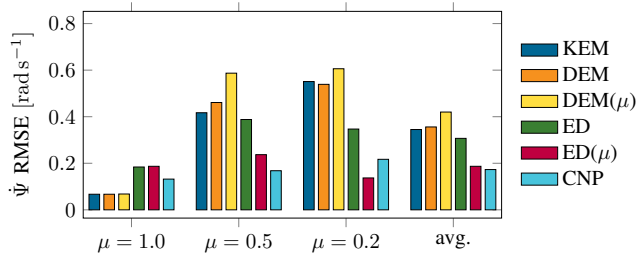
\begin{figure}[thpb]

	\input{figures/04_Evaluation_Scenario}
	\caption{Prediction evaluation under varied scenarios. Results separated according to different road (friction) conditions.}
	\label{fig:scenario}
\end{figure}

Unlike previous evaluations, this evaluation is based on scenarios that are not part of the training. Two urban scenarios, two interurban driving scenarios, two racetrack scenarios (Nordschleife Nürburgring, Hockenheimring) and one mountain pass scenario serve as the test cases. In all scenarios, velocity is varied appropriately. Friction parameters are likewise varied to simulate dry, wet and icy road surfaces. The numerical evaluation in Fig. \ref{fig:scenario} illustrates a similar pattern as in the previous evaluations. While CNP receives only the acceleration as an additional input compared to the simplest KST model, the error is comparable to complex physical models, which furthermore have the true parameters given. To complete the analysis, Fig. \ref{fig_trajectory} shows the prediction of the yaw rate for various friction coefficients for the Hockenheimring racetrack scenario. In the dry road domain, the CNP performs lowest while showing the highest uncertainty. In the area of wet and icy roads, the prediction is better while the uncertainty is quite low. The scenario on the icy road is significantly shorter, since the same driving style over different road parameters causes the vehicle to leave the track in this icy condition. 

\begin{figure}[t!]
		\input{figures/04_Evaluation_Scenario_Qualtiative}
		\caption{Icy road condtions ($\mu=0.2$).}
	\end{subfigure}
	\caption{Qualitative trajectory evaluation over (a) dry, (b) wet and (c) icy roads on the Hockenheim racetrack. The vertical dashed line separates the context set on the left from the predictions on the right. The time horizon in (c) is shorter since, given the same driving behavior, the vehicle leaves the road under icy conditions.}
	\label{fig_trajectory}
	
\end{figure}

\subsection{Vehicle variation}
The training and evaluation up to now is based on the use of a VW Beatle, the CarMaker DemoCar. In this evaluation the transferability of the CNP to other vehicles is evaluated. Here, the evaluation from Section \ref{Scenario} is applied to a small car (Honda Fit), an SUV (BMW X5), a van (VW T6) and a sports car (Porsche 911). The results for dry roads (Tab. \ref{tab:dry}), wet roads (Tab. \ref{tab:wet}), icy roads (Tab. \ref{tab:icy}) as well as average (Tab. \ref{tab:all}) demonstrate that the CNP performs similarly on other vehicles. This shows that the CNP is able to extract the characteristics of the present dynamics from a context dataset and to transfer learned knowledge to a large extent.

%% file: figures/04_Evaluation_Friction.tex
\begin{tikzpicture}[ scale=0.8] 
	\begin{axis}[
		ybar,
		ymax=0.75,
		legend image code/.code={
			\draw [#1] (0cm,-0.1cm) rectangle (0.6cm,0.1cm);
		},
		height= 5.0cm,
		width=8.5cm,
		bar width=5pt,
		enlargelimits=0.15,
		symbolic x coords={$\mu=0.75$,$\mu=0.35$,$\mu=0.1$,avg.},
		xtick={$\mu=0.75$,$\mu=0.35$,$\mu=0.1$,avg.},
		ylabel={$\dot{\Psi}$ RMSE $[\SI{}{\radian\per\second}]$},
		y,
		legend style={
			overlay,        
			at={(1.03,0.5)},
			anchor=west,
			draw=none,
		},
		legend cell align={left},
		]
		
		typeset ticklabels with strut,
		legend pos=outer north east,
		legend cell align=left,
		smooth,
		
		\addplot[fill=MidnightBlue,] coordinates{($\mu=0.75$,0.148) ($\mu=0.35$,0.465) ($\mu=0.1$,0.736) (avg.,0.442)}; 
		\addlegendentry{KST};
		\addplot[fill=BurntOrange] coordinates{($\mu=0.75$,0.157) ($\mu=0.35$,0.460) ($\mu=0.1$,0.728) (avg.,0.441)};  
		\addlegendentry{DST};
		\addplot[fill=Goldenrod] coordinates{($\mu=0.75$,0.170) ($\mu=0.35$,0.479) ($\mu=0.1$,0.621) (avg.,0.416)}; 
		\addlegendentry{DST($\mu$)}
		\addplot[fill=OliveGreen] coordinates{($\mu=0.75$,0.167) ($\mu=0.35$,0.269) ($\mu=0.1$,0.303) (avg.,0.365)};    
		\addlegendentry{STD};
		\addplot[fill=purple] coordinates{($\mu=0.75$,0.151) ($\mu=0.35$,0.143) ($\mu=0.1$,0.082) (avg.,0.146)}; 
		\addlegendentry{STD($\mu$)}
		\addplot[fill=SkyBlue] coordinates{($\mu=0.75$,0.202) ($\mu=0.35$,0.189) ($\mu=0.1$,0.215) (avg.,0.202)}; 
		\addlegendentry{CNP};


	\end{axis} 
\end{tikzpicture}

%% file: figures/04_Evaluation_Mass.tex
\begin{tikzpicture}[ scale=0.8]
	\begin{axis}[
		ybar,
		ymax=0.75,
		legend image code/.code={
			\draw [#1] (0cm,-0.1cm) rectangle (0.6cm,0.1cm);
		},
		height= 5.0cm,
		width=8.5cm,
		bar width=5pt,
		enlargelimits=0.15,
		symbolic x coords={$\mu=1.0$,$\mu=0.5$,$\mu=0.2$,avg.},
		xtick={$\mu=1.0$,$\mu=0.5$,$\mu=0.2$,avg.},
		ylabel={$\dot{\Psi}$  RMSE $[\SI{}{\radian\per\second}]$},
		y,
		legend style={
			overlay,        
			at={(1.03,0.5)},
			anchor=west,
			draw=none,
		},
		legend cell align={left},
		]
		
		typeset ticklabels with strut,
		legend pos=outer north east,
		legend cell align=left,
		smooth,
		
		\addplot[fill=MidnightBlue,] coordinates{($\mu=1.0$,0.082) ($\mu=0.5$,0.333) ($\mu=0.2$,0.612) (avg.,0.314)}; 
		\addlegendentry{KST};
		\addplot[fill=BurntOrange] coordinates{($\mu=1.0$,0.080) ($\mu=0.5$,0.353) ($\mu=0.2$,0.582) (avg.,0.337)};  
		\addlegendentry{DST};
		\addplot[fill=Goldenrod] coordinates{($\mu=1.0$,0.081) ($\mu=0.5$,0.413) ($\mu=0.2$,0.593) (avg.,0.362)}; 
		\addlegendentry{DST($\mu,m$)}
		\addplot[fill=OliveGreen] coordinates{($\mu=1.0$,0.126) ($\mu=0.5$,0.228) ($\mu=0.2$,0.284) (avg.,0.212)};    
		\addlegendentry{STD};
		\addplot[fill=purple] coordinates{($\mu=1.0$,0.133) ($\mu=0.5$,0.177) ($\mu=0.2$,0.096) (avg.,0.135)}; 
		\addlegendentry{STD($\mu,m$)}
		\addplot[fill=SkyBlue] coordinates{($\mu=1.0$,0.183) ($\mu=0.5$,0.196) ($\mu=0.2$,0.202) (avg.,0.194)}; 
		\addlegendentry{CNP};


	\end{axis} 
\end{tikzpicture}

%% file: figures/04_Evaluation_Scenario.tex
\begin{tikzpicture}[ scale=0.8] 
	\begin{axis}[
		ybar,
		ymax=0.75,
		legend image code/.code={
			\draw [#1] (0cm,-0.1cm) rectangle (0.6cm,0.1cm);
		},
		height= 5.0cm,
		width=8.5cm,
		bar width=5pt,
		enlargelimits=0.15,
		symbolic x coords={$\mu=1.0$,$\mu=0.5$,$\mu=0.2$,avg.},
		xtick={$\mu=1.0$,$\mu=0.5$,$\mu=0.2$,avg.},
		ylabel={$\dot{\Psi}$ RMSE $[\SI{}{\radian\per\second}]$},
		y,
		legend style={
			overlay,        
			at={(1.03,0.5)},
			anchor=west,
			draw=none,
		},
		legend cell align={left},
		]
		
		typeset ticklabels with strut,
		legend pos=outer north east,
		legend cell align=left,
		smooth,
		
		\addplot[fill=MidnightBlue,] coordinates{($\mu=1.0$,0.067) ($\mu=0.5$,0.417) ($\mu=0.2$,0.551) (avg.,0.345)}; 
		\addlegendentry{KEM};
		\addplot[fill=BurntOrange] coordinates{($\mu=1.0$,0.067) ($\mu=0.5$,0.461) ($\mu=0.2$,0.539) (avg.,0.356)};  
		\addlegendentry{DEM};
		\addplot[fill=Goldenrod] coordinates{($\mu=1.0$,0.068) ($\mu=0.5$,0.587) ($\mu=0.2$,0.606) (avg.,0.420)}; 
		\addlegendentry{DEM($\mu$)}
		\addplot[fill=OliveGreen] coordinates{($\mu=1.0$,0.184) ($\mu=0.5$,0.388) ($\mu=0.2$,0.347) (avg.,0.307)};    
		\addlegendentry{ED};
		\addplot[fill=purple] coordinates{($\mu=1.0$,0.187) ($\mu=0.5$,0.237) ($\mu=0.2$,0.137) (avg.,0.187)}; 
		\addlegendentry{ED($\mu$)}
		\addplot[fill=SkyBlue] coordinates{($\mu=1.0$,0.132) ($\mu=0.5$,0.168) ($\mu=0.2$,0.217) (avg.,0.173)}; 
		\addlegendentry{CNP};
		
		
		
	\end{axis} 
\end{tikzpicture}

%% file: figures/04_Evaluation_Scenario_Qualtiative.tex
\def \file {notesdata/data/data0.csv}
\def \xmin {0}
\def \xmax {200}
\def \ymin {-0.5}
\def \ymax {+0.5}

\begin{subfigure}[thpb]{\columnwidth}
	\centering
	\begin{tikzpicture}[scale=1, font=\footnotesize]
		\begin{axis}[
			width=0.97\columnwidth,
			height=4.6cm,
			xmin=\xmin,
			xmax=\xmax,
			ymin=\ymin,
			ymax=\ymax,
			xlabel={\( t \) [s]},
			ylabel={$\dot{\Psi}$ RMSE $[\SI{}{\radian\per\second}]$},
			legend style={at={(0.5,+1.12)},anchor=south, legend columns=3, column sep=0.29cm, legend cell align=left},,
			xmajorgrids,
			ymajorgrids
			]
			
			\addplot[black, thick, line join=bevel, domain=0:30] table [col sep=comma, x index=1, y index=2] {\file}; 
			\addplot[MidnightBlue, very thick, densely dotted, line join=bevel, restrict x to domain=50:200] table [col sep=comma, x index=1, y index=3] {\file}; 
			\addplot[Goldenrod, very thick, dashed, line join=bevel, restrict x to domain=50:200] table [col sep=comma, x index=1, y index=4] {\file};
			\addplot[purple,very thick, dash dot, line join=bevel, restrict x to domain=50:200] table [col sep=comma, x index=1, y index=5]  {\file};
			\addplot[SkyBlue, very thick, dotted, line join=bevel, restrict x to domain=50:200] table [col sep=comma, x index=1, y index=6]  {\file};
			\addplot[teal!30, very thick, name path = A, draw opacity=0.0, restrict x to domain=50:200] table [col sep=comma, x index=1, y index=7]  {\file};
			\addplot[teal!30, very thick, name path = B, draw opacity=0.0, restrict x to domain=50:200] table [col sep=comma, x index=1, y index=8]  {\file};
			\addplot [teal!30, very thick, fill opacity=0.6, restrict x to domain=50:200] fill between [of = A and B,];
			\draw [gray, very thick, dashed] (50,2) -- (50,-2);
			
			\legend{
				CM,
				KST,
				DST($\mu$),
				STD($\mu$),
				CNP:$\mu$,
				,
				,
				CNP:$2\sigma^{2}$,
				,
			}

		\end{axis}
	\end{tikzpicture}
	\captionsetup{justification=centering}
	\caption{Dry road condtions ($\mu=1.0$).}
\end{subfigure}

\bigskip

\def \file {notesdata/data/data1.csv}
\def \xmin {0}
\def \xmax {200}
\def \ymin {-3}
\def \ymax {+3}

\begin{subfigure}[b]{\columnwidth}
	\begin{tikzpicture}[scale=1,font=\footnotesize]
		\begin{axis}[
			width=\columnwidth,
			height=4.6cm,
			xmin=\xmin,
			xmax=\xmax,
			ymin=\ymin,
			ymax=\ymax,
			xlabel={\( t \) [s]},
			ylabel={$\dot{\Psi}$ RMSE $[\SI{}{\radian\per\second}]$},
			legend style={legend columns=3,at={(1.42,1.19)}},
			legend style={legend columns=1, font=\footnotesize},
			xmajorgrids,
			ymajorgrids
			]
			
			\addplot[black, thick, line join=bevel, domain=0:30] table [col sep=comma, x index=1, y index=2] {\file}; 
			\addplot[MidnightBlue, very thick, densely dotted, line join=bevel, restrict x to domain=50:200] table [col sep=comma, x index=1, y index=3] {\file}; 
			\addplot[Goldenrod,very thick, dashed, line join=bevel, restrict x to domain=50:200] table [col sep=comma, x index=1, y index=4] {\file};
			\addplot[purple,very thick, dash dot, line join=bevel, restrict x to domain=50:200] table [col sep=comma, x index=1, y index=5]  {\file};
			\addplot[SkyBlue, very thick, dotted, line join=bevel, restrict x to domain=50:200] table [col sep=comma, x index=1, y index=6]  {\file};
			\addplot[teal!30, very thick, name path = A, draw opacity=0.0, restrict x to domain=50:200] table [col sep=comma, x index=1, y index=7]  {\file};
			\addplot[teal!30, very thick, name path = B, draw opacity=0.0, restrict x to domain=50:200] table [col sep=comma, x index=1, y index=8]  {\file};
			\addplot [teal!30, very thick, fill opacity=0.6, restrict x to domain=50:200] fill between [of = A and B,];
			\draw [gray, very thick, dashed] (50,3) -- (50,-3);
			

		\end{axis}
	\end{tikzpicture}
	\caption{Wet road condtions ($\mu=0.5$).}
\end{subfigure}

\bigskip

\def \file {notesdata/data/data2.csv}
\def \xmin {0}
\def \xmax {27.6}
\def \ymin {-4}
\def \ymax {+1}

\begin{subfigure}[b]{\columnwidth}
	\begin{tikzpicture}[scale=1,font=\footnotesize]
		\begin{axis}[
			width=\columnwidth,
			height=4.6cm,
			xmin=\xmin,
			xmax=\xmax,
			ymin=\ymin,
			ymax=\ymax,
			xlabel={\( t \) [s]},
			ylabel={$\dot{\Psi}$ RMSE $[\SI{}{\radian\per\second}]$},
			legend style={at={(0.5,-0.3)},anchor=north, legend columns=3, font=\footnotesize},
			xmajorgrids,
			ymajorgrids
			]
			
			\addplot[black, thick, line join=bevel, domain=0:30] table [col sep=comma, x index=1, y index=2] {\file}; 
			\addplot[MidnightBlue, very thick, densely dotted, line join=bevel, restrict x to domain=2.76:30] table [col sep=comma, x index=1, y index=3] {\file}; 
			\addplot[Goldenrod,very thick, dashed, line join=bevel, restrict x to domain=2.76:30] table [col sep=comma, x index=1, y index=4] {\file};
			\addplot[purple,very thick, dash dot, line join=bevel, restrict x to domain=2.76:30] table [col sep=comma, x index=1, y index=5]  {\file};
			\addplot[SkyBlue, very thick, dotted, line join=bevel, restrict x to domain=2.75:30] table [col sep=comma, x index=1, y index=6]  {\file};
			\addplot[teal!30, very thick, name path = A, draw opacity=0.0, restrict x to domain=2.76:30] table [col sep=comma, x index=1, y index=7]  {\file};
			\addplot[teal!30, very thick, name path = B, draw opacity=0.0, restrict x to domain=2.76:30] table [col sep=comma, x index=1, y index=8]  {\file};
			\addplot [teal!30, very thick, fill opacity=0.6, restrict x to domain=2.76:30] fill between [of = A and B,];
			\draw [gray, very thick, dashed] (27.6*0.1,2) -- (27.6*0.1,-5);
			
			%

		\end{axis}
	\end{tikzpicture}

%% file: src/05_Conclusion.tex
\section{CONCLUSIONS AND FUTURE WORKS}

The investigation of the meta-learning approach conditional neural processes in terms of yaw rate prediction shows that the methodology is robust to environmental and operational variations and can even be transferred to other cars. It is also demonstrated that CNP offer a low mean prediction error while simultaneously maintaining a low computational complexity. Thus, a kind of model-order reduction of classical models is feasible. Moreover, the methodology also provides the possibility of uncertainty quantification, and more recent architectures of the neural process family are proven to yield improvement in this area. Future work remains to investigate the performance difference towards more advanced neural processes models, especially with respect to variance, which in general needs to be investigated in more detail. In addition, the impact of stochastic measurement inputs remains to be investigated. Ultimately, the behavior of a resulting stochastic hybrid vehicle model for trajectory planning in autonomous driving is to be investigated. While the methodology offers great advantages, and resolves e.g. out-of-distribution issues in a targeted manner, the safety assurance of AI will be a key research question.

\begin{table}[ht]
	\caption{RMSE Vehicle variation dry roads.}
	\begin{center}
		\begin{tabular}{|c|p{0.039\textwidth}|p{0.039\textwidth}|p{0.039\textwidth}|p{0.039\textwidth}|p{0.039\textwidth}|p{0.039\textwidth}|}
			\hline
			Vehicle&KST&DST& DST($\mu$)&STD&STD($\mu$)&CNP\\
			\hline
			\hline
			Honda Fit &\textbf{0,112}&	0,154&	0,159&	0,196&	0,205&	0,166\\
			\hline
			VW Beatle &\textbf{0,067}&	\textbf{0,067}&	0,068&	0,184&	0,187&	0,132 \\
			\hline
			BMW X5 &0,107&	0,172&	0,098&	0,253&	\textbf{0,140}&	0,150 \\
			\hline
			VW T6 &0,062&	0,059&	\textbf{0,028}&	0,136&	0,120&	0,100 \\
			\hline
			Porsche 911&\textbf{0,054}&	0,077&	0,071&	0,201&	0,151&	0,142 \\
			\hline
		\end{tabular}
		\label{tab:dry}
	\end{center}
\end{table}


\begin{table}[ht]
	\caption{RMSE Vehicle variation wet roads.}
	\begin{center}
		\begin{tabular}{|c|p{0.039\textwidth}|p{0.039\textwidth}|p{0.039\textwidth}|p{0.039\textwidth}|p{0.039\textwidth}|p{0.039\textwidth}|}
			\hline
			Vehicle & KST & DST & DST($\mu$) & STD & STD($\mu$) & CNP \\
			\hline
			\hline
			Honda Fit &0,305	&0,396	&0,637&	0,423&	0,289&	\textbf{0,178} \\
			\hline
			VW Beatle &0,417	&0,461	&0,587&	0,388&	0,237&	\textbf{0,168} \\
			\hline
			BMW X5 &0,226	&0,296	&0,283&	0,334&	\textbf{0,165}&	0,191 \\
			\hline
			VW T6 &0,130	&0,138	&0,126&	0,151&	0,125&	\textbf{0,114} \\
			\hline
			Porsche 911&0,306	&0,364	&0,440&	0,644&	0,331&	\textbf{0,190} \\
			\hline
		\end{tabular}
		\label{tab:wet}
	\end{center}
\end{table}

\begin{table}[ht]
	\caption{RMSE Vehicle variation icy roads.}
	\begin{center}
		\begin{tabular}{|c|p{0.039\textwidth}|p{0.039\textwidth}|p{0.039\textwidth}|p{0.039\textwidth}|p{0.039\textwidth}|p{0.039\textwidth}|}
			\hline
			Vehicle & KST & DST & DST($\mu$) & STD & STD($\mu$) & CNP \\
			\hline
			\hline
			Honda Fit &0,660&	0,620&	0,657&	0,387&	\textbf{0,251}&	0,260 \\
			\hline
			VW Beatle &0,551&	0,539&	0,606&	0,347&	\textbf{0,137}&	0,217\\
			\hline
			BMW X5 &0,545&	0,536&	0,477&	0,422&	0,401&	\textbf{0,284} \\
			\hline
			VW T6 &0,348&	0,339&	0,280&	0,325&	0,244&	\textbf{0,179} \\
			\hline
			Porsche 911&0,746&	0,781&	0,785&	0,472&	\textbf{0,254}&	0,356\\
			\hline
		\end{tabular}
		\label{tab:icy}
	\end{center}
\end{table}

\begin{table}[ht]
	\caption{RMSE Vehicle variation on average.}
	\begin{center}
		\begin{tabular}{|c|p{0.039\textwidth}|p{0.039\textwidth}|p{0.039\textwidth}|p{0.039\textwidth}|p{0.039\textwidth}|p{0.039\textwidth}|}
			\hline
			Vehicle & KST & DST & DST($\mu$) & STD & STD($\mu$) & CNP \\
			\hline
			\hline
			Honda Fit &0,359&	0,390&	0,484&	0,336&	0,248&	\textbf{0,201} \\
			\hline
			VW Beatle &0,345&	0,356&	0,420&	0,307&	0,187&	\textbf{0,173} \\
			\hline
			BMW X5 &0,292&	0,335&	0,286&	0,336&	0,235&	\textbf{0,208} \\
			\hline
			VW T6 &0,180&	0,179&	0,145&	0,204&	0,163&	\textbf{0,131} \\
			\hline
			Porsche 911&0,369&	0,407&	0,432&	0,439&	0,245&	\textbf{0,229} \\
			\hline
		\end{tabular}
		\label{tab:all}
	\end{center}
\end{table}